\documentclass[conference]{IEEEtran}
\IEEEoverridecommandlockouts

\usepackage{cite}
\usepackage{amsmath,amssymb,amsfonts}
\usepackage{algorithmic}
\usepackage{graphicx}
\usepackage{textcomp}
\usepackage{xcolor}
\usepackage{url}
\usepackage{booktabs}
\usepackage{balance}
\def\BibTeX{{\rm B\kern-.05em{\sc i\kern-.025em b}\kern-.08em
    T\kern-.1667em\lower.7ex\hbox{E}\kern-.125emX}}
\newcommand{\etal}{et al.}
\begin{document}

\title{HazeCLIP: Towards Language Guided Real-World Image Dehazing
\thanks{$^*$Equal contribution. $^\dagger$Corresponding author.}
\thanks{The work was supported in part by the National Natural Science Foundation of China under Grant 62301310, and in part by Sichuan Science and Technology Program under Grant 2024NSFSC1426.}
}

\author{\IEEEauthorblockN{Ruiyi Wang$^*$, Wenhao Li$^*$, Xiaohong Liu$^\dagger$, Chunyi Li, Zicheng Zhang, Xiongkuo Min, and Guangtao Zhai}
\IEEEauthorblockA{
\textit{Shanghai Jiao Tong University, Shanghai, China}
 \\
{\{thomas25, sevenhao, xiaohongliu, lcysyzxdxc, zzc1998, minxiongkuo, zhaiguangtao\}@sjtu.edu.cn}}
}

\maketitle

\begin{abstract}
Existing methods have achieved remarkable performance in image dehazing, particularly on synthetic datasets. However, they often struggle with real-world hazy images due to domain shift, limiting their practical applicability. This paper introduces HazeCLIP, a language-guided adaptation framework designed to enhance the real-world performance of pre-trained dehazing networks. Inspired by the Contrastive Language-Image Pre-training (CLIP) model's ability to distinguish between hazy and clean images, we leverage it to evaluate dehazing results. Combined with a region-specific dehazing technique and tailored prompt sets, the CLIP model accurately identifies hazy areas, providing a high-quality, human-like prior that guides the fine-tuning process of pre-trained networks. Extensive experiments demonstrate that HazeCLIP achieves state-of-the-art performance in real-word image dehazing, evaluated through both visual quality and image quality assessment metrics. Codes are available at \color{magenta}{\url{https://github.com/Troivyn/HazeCLIP}}.
\end{abstract}
 

\section{Introduction}

Single image dehazing aims to restore clean images from hazy ones that suffer from reduced contrast and limited visibility. Similar to other restoration tasks~\cite{restoration1, restoration2, glare}, this challenging task is crucial for high-level vision applications such as object detection~\cite{DENet_detection, DSNet_detection} and semantic scene understanding~\cite{scene, scene_understanding, scene2}, making it a longstanding problem. Early dehazing algorithms~\cite{prior1, Fattal, prior2, prior3, prior4} typically estimate parameters of the atmospheric scattering model~\cite{asm2} using statistical priors. For example, He~\etal~\cite{prior3} achieved impressive dehazing results using the Dark Channel Prior (DCP). However, while these prior-based models can perform well without training, their performance is often limited and fragile because hand-crafted priors cannot adapt to the diversity of real-world images.

With the availability of synthetic datasets and Convolutional Neural Networks (CNNs), numerous learning-based approaches have emerged~\cite{GDN, gdn+, Dong_2020_CVPR, DehazeDCT, icassp1, icassp2, addref1, addref2}. Particularly, Liu~\etal~\cite{GDN} proposed an attention-based multi-scale grid network for image dehazing. However, due to the significant domain gap, their performance drops dramatically when applied to real-world hazy images. To overcome this challenge, several works have been proposed for real-world image dehazing~\cite{hazinganddehazing, Shao_2020_CVPR, Chen_2021_CVPR, Wu_2023_CVPR}. Among them, many studies reintroduce prior knowledge. For example, Chen~\etal~\cite{Chen_2021_CVPR} employed three statistical image priors for unsupervised fine-tuning, while Wu~\etal~\cite{Wu_2023_CVPR} leveraged latent discrete  priors in pre-trained VQGAN~\cite{VQGAN}. However, directly using handcrafted or learned priors cannot avoid the inherent flaws of prior-based methods.
\begin{figure}[t]
	\centering
	\includegraphics[width=0.98\linewidth]{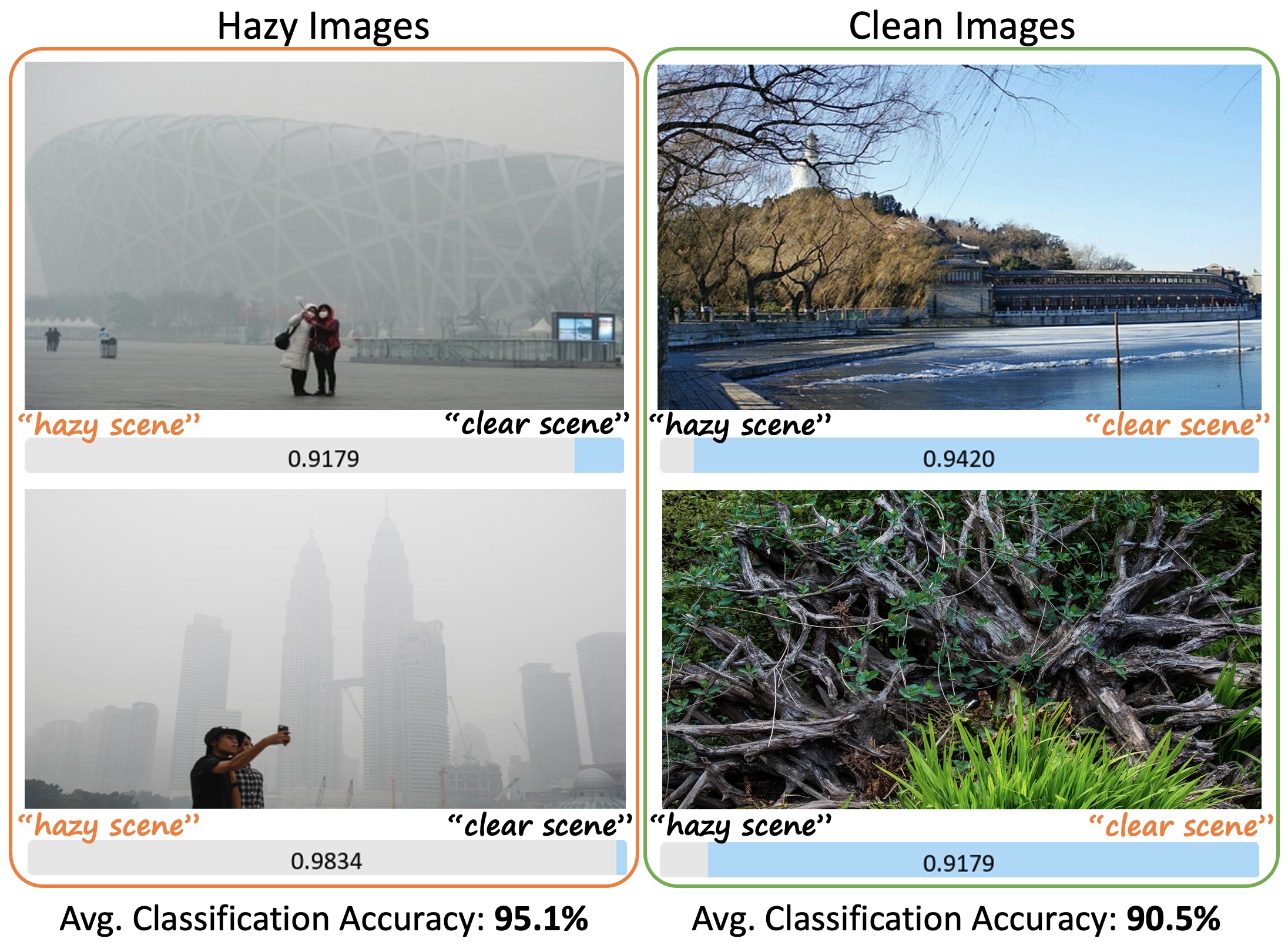}
    \caption{CLIP model is capable of distinguishing between hazy and clean images. Classification probabilities for example images and average accuracy for hazy and clean image sets are reported.}
	\label{fig:motivation}
\end{figure}

Recently, a few methods have made breakthroughs in low-level vision tasks leveraging text information, including~\cite{coser, promptsr, language1, language2}. Specifically, Luo~\etal~\cite{luo2023controlling} proposed a method to control the CLIP model to disentangle degradation factors from the image features. Zhang~\etal~\cite{CLIP-LIT} proposed an iterative prompt learning method for backlit image enhancement. Considering the success of these methods and the flexibility of natural language compared to statistical priors, we decided to leverage the power of CLIP model for image dehazing.

\begin{figure*}[t]
  \centering
  \includegraphics[width=1.00\linewidth]{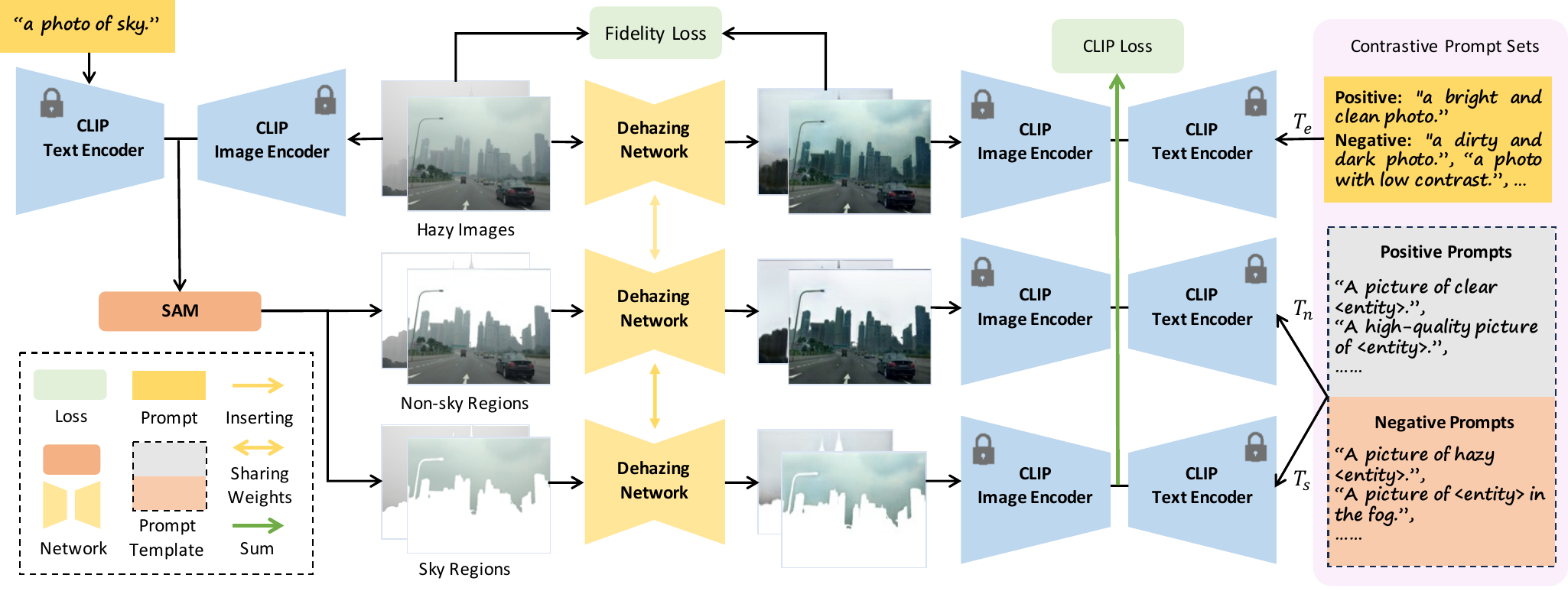}
  \caption{Overview of the proposed HazeCLIP framework. Real-world hazy images are first separated into sky and non-sky regions using Segment Anything Model (SAM). Combined with the CLIP model, three contrastive prompt sets are applied to guide the adaptation process. The enhancing prompt set aims to improve overall image quality, and non-sky and sky dehazing prompt sets specifically guide dehazing in their respective regions.}
  \label{fig:method}
\end{figure*}

To overcome the drawbacks of existing dehazing methods, we present HazeCLIP, a language-guided adaptation framework for real image dehazing that generalizes dehazing networks to real-world domain. Inspired by CLIP model's ability to distinguish between hazy and clean images (Fig.~\ref{fig:motivation}), our approach leverages its rich visual-language prior instead of relying on the scattering model or statistical priors. Since CLIP model is trained on a diverse set of images, its prior is more robust than traditional statistical ones. Additionally, CLIP model's perception aligns closely with human perception, allowing our method to generate images with superior visual quality. Nevertheless, potentially due to training data bias where images with haze captions often feature grey skies, CLIP model yields excessively high haze similarity scores for sky areas, impeding its ability to guide haze removal in other areas. To overcome this challenge, we propose a region-specific dehazing technique that separately processes the sky and non-sky regions during fine-tuning. This approach improves CLIP model's accuracy in identifying hazy areas, providing more effective guidance for haze removal in diverse scenes.


We summarize our contributions as follows:

 $\diamond$ We propose a general language-guided adapting framework for real-world image dehazing. This framework can be easily combined with many existing dehazing networks without modifying the network architecture.
 
$\diamond$ We are pioneers to leverage the power of vision-language models in image dehazing. By designing contrastive prompt sets, the CLIP similarity score can guide the fine-tuning of the dehazing network. We also propose a region-specific dehazing technique to resolve the issue where CLIP model fails to accurately detect hazy regions.

$\diamond$ Extensive experiments demonstrate that our HazeCLIP achieves the SOTA real-world dehazing performance. 
\section{Proposed Method}

\subsection{Overview}
As shown in Fig.~\ref{fig:method}, Our HazeCLIP is a fine-tuning framework that guides the pre-trained dehazing network towards real domain leveraging a frozen CLIP model. First we adopt a dehazing network $\mathcal{M}$ pre-trained with synthetic data as the backbone. Given that our approach constitutes a general framework, there are no specific requirements for the chosen dehazing network. Observing the random language-image similarity maps generated by the CLIP model, we employ CLIP surgery and a region-specific dehazing technique to achieve accurate haze detection. To enhance language guidance, we develop three contrastive prompt sets, denoted as $T_e, T_n, T_s$. We provide further details on the key components of our framework below.

\begin{figure}[t]
	\centering
    \begin{minipage}[h]{0.24\linewidth}
		\centering
		\includegraphics[width=\linewidth]{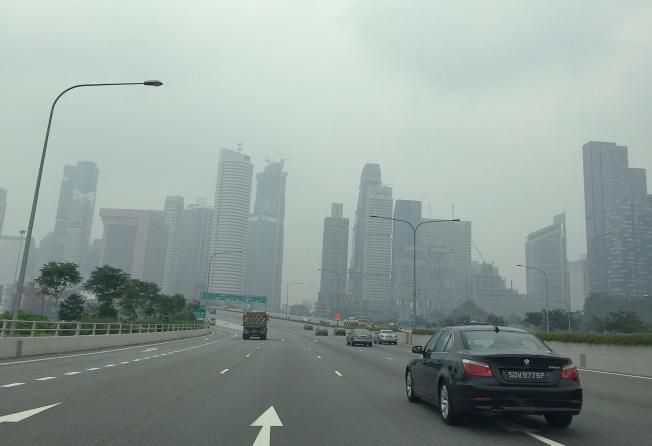}
	\end{minipage}
	\begin{minipage}[h]{0.24\linewidth}
		\centering
		\includegraphics[width=\linewidth]{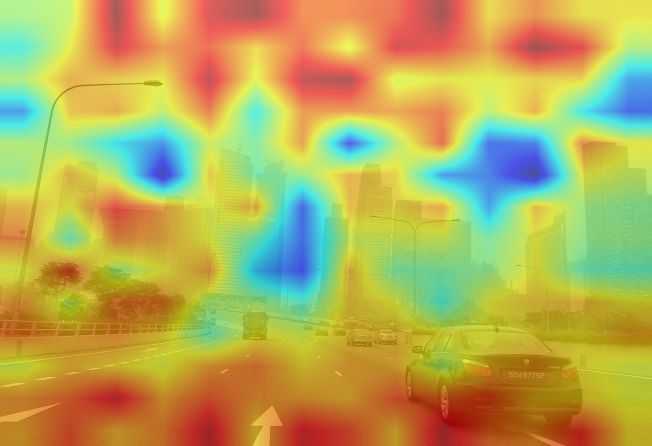}
	\end{minipage}
	\begin{minipage}[h]{0.24\linewidth}
		\centering
		\includegraphics[width=\linewidth]{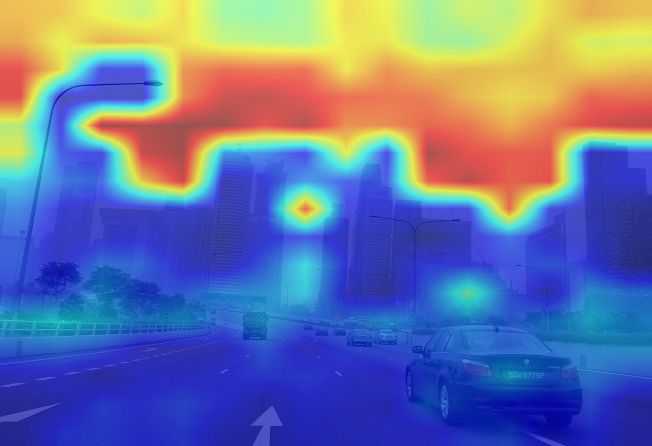}
	\end{minipage}
	\begin{minipage}[h]{0.24\linewidth}
		\centering
		\includegraphics[width=\linewidth]{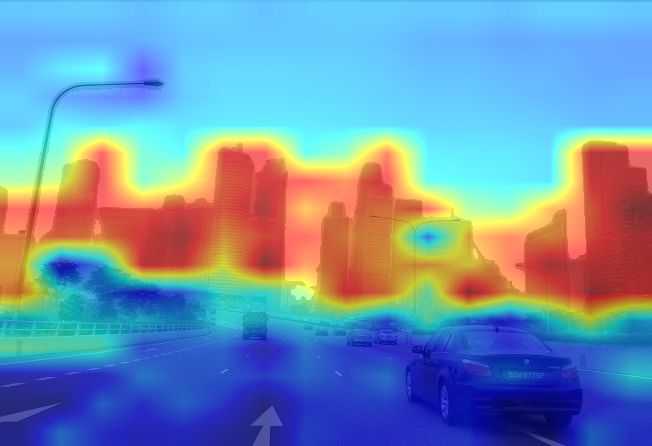}
	\end{minipage}

    \begin{minipage}[h]{0.24\linewidth}
		\centering
		\includegraphics[width=\linewidth]{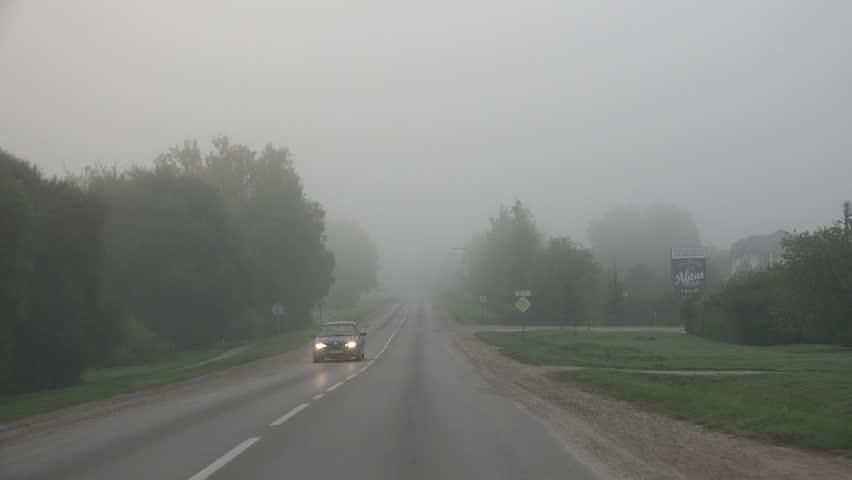}
            \small{(a)}
	\end{minipage}
	\begin{minipage}[h]{0.24\linewidth}
		\centering
		\includegraphics[width=\linewidth]{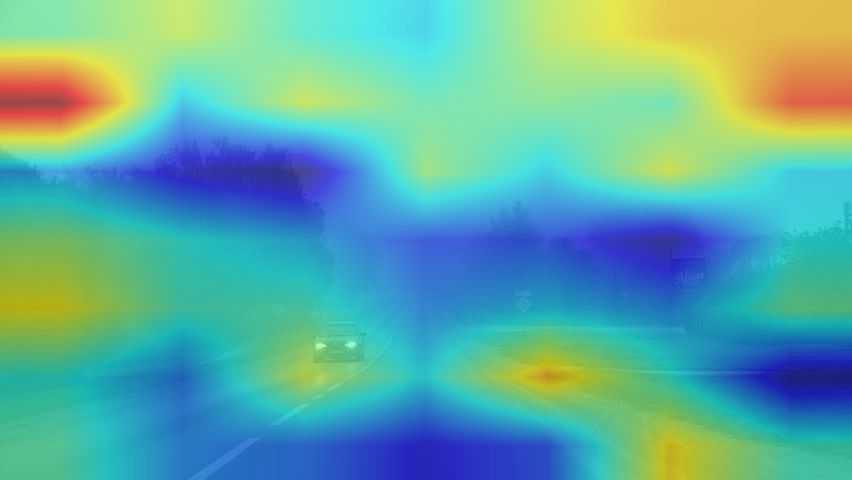}
            \small{(b)}
	\end{minipage}
	\begin{minipage}[h]{0.24\linewidth}
		\centering
		\includegraphics[width=\linewidth]{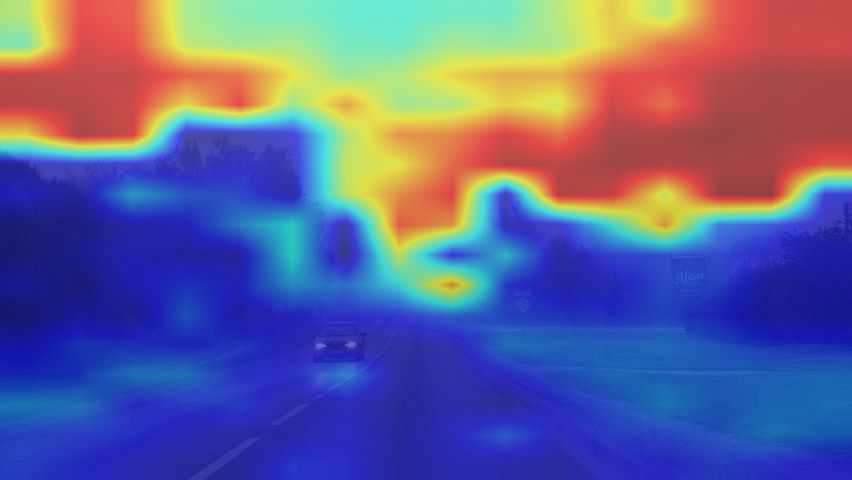}
            \small{(c)}
	\end{minipage}
	\begin{minipage}[h]{0.24\linewidth}
		\centering
		\includegraphics[width=\linewidth]{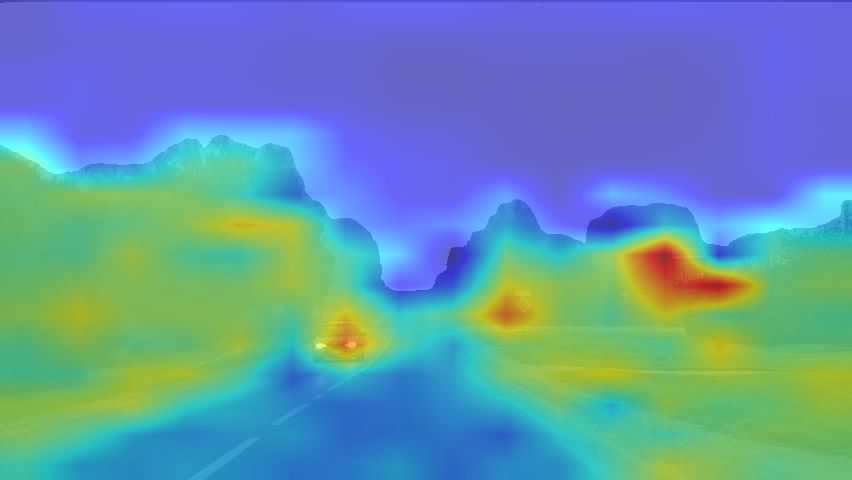}
            \small{(d)}
	\end{minipage}

	\caption{Language-image similarity maps for hazy images. By removing the sky, the CLIP model can focus more effectively on scene dehazing.
(a) Hazy images, (b) Raw similarity maps, (c) Maps of CLIP surgery~\cite{clip_surgery}, (d) Maps after sky masking.}
	\label{fig:calibration}
\end{figure}

\subsection{Region-Specific Dehazing}
In Fig.~\ref{fig:calibration}, we present rough language-image similarity maps for hazy images, calculated by measuring the cosine similarity between the features of image patches and description of hazy images. While CLIP model effectively classifies hazy and clean images, it struggles to accurately locate hazy regions, resulting in random similarity maps. To address this limitation, we employ CLIP surgery~\cite{clip_surgery}, a technique designed to enhance CLIP model's explainability. However, a significant issue arises as sky region dominates the similarity. In consequence, when CLIP model evaluates the entire image, haze residuals in non-sky regions, such as hazy buildings, are overlooked, leading to sub-optimal dehazing performance.

\begin{figure*}[t]
	\centering
    \begin{minipage}[h]{0.119\linewidth}
		\centering
		\includegraphics[width=\linewidth]{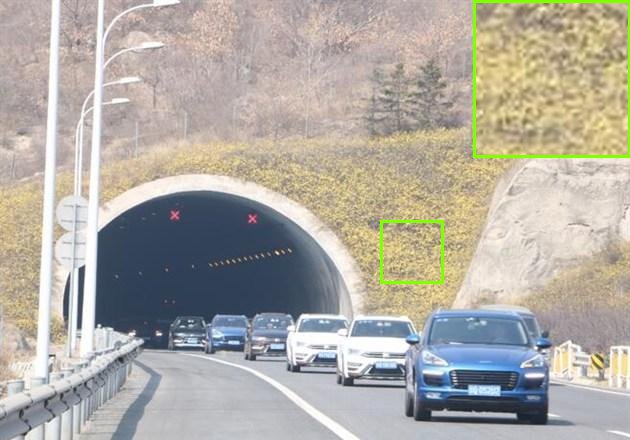}
	\end{minipage}
	\begin{minipage}[h]{0.119\linewidth}
		\centering
		\includegraphics[width=\linewidth]{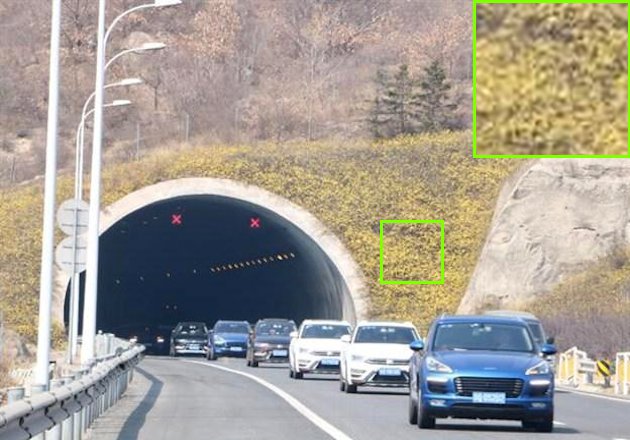}
	\end{minipage}
	\begin{minipage}[h]{0.119\linewidth}
		\centering
		\includegraphics[width=\linewidth]{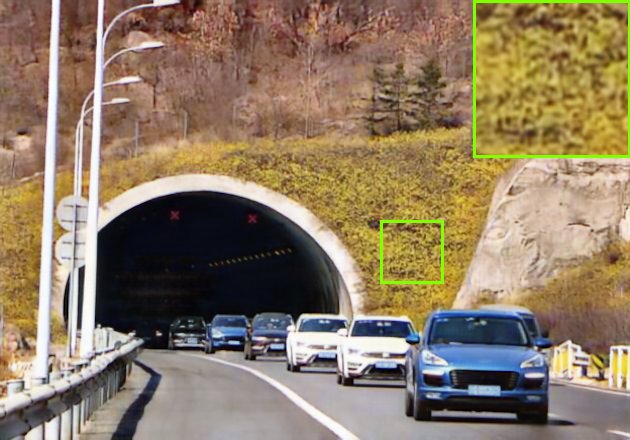}
	\end{minipage}
	\begin{minipage}[h]{0.119\linewidth}
		\centering
		\includegraphics[width=\linewidth]{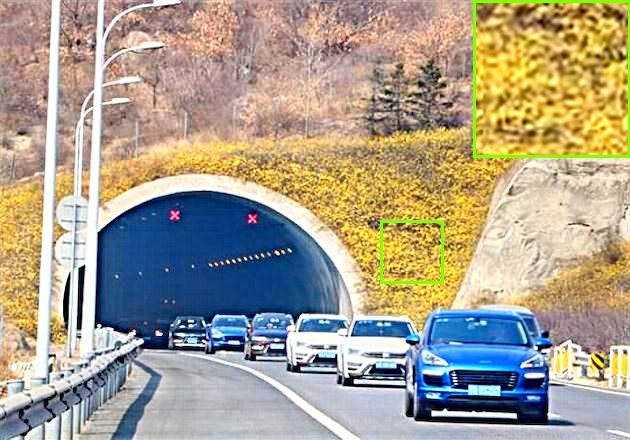}
	\end{minipage}
	\begin{minipage}[h]{0.119\linewidth}
		\centering
		\includegraphics[width=\linewidth]{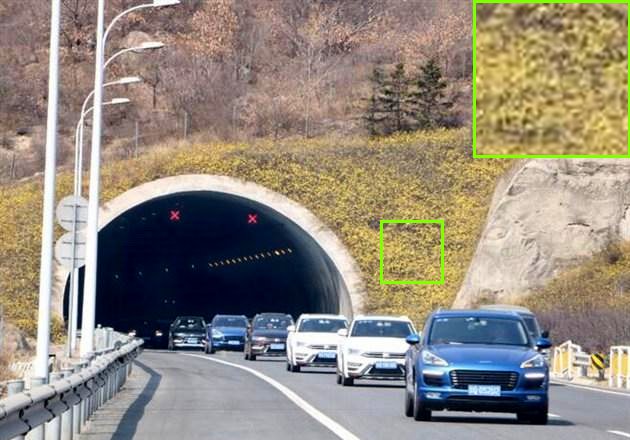}
	\end{minipage}
	\begin{minipage}[h]{0.119\linewidth}
		\centering
		\includegraphics[width=\linewidth]{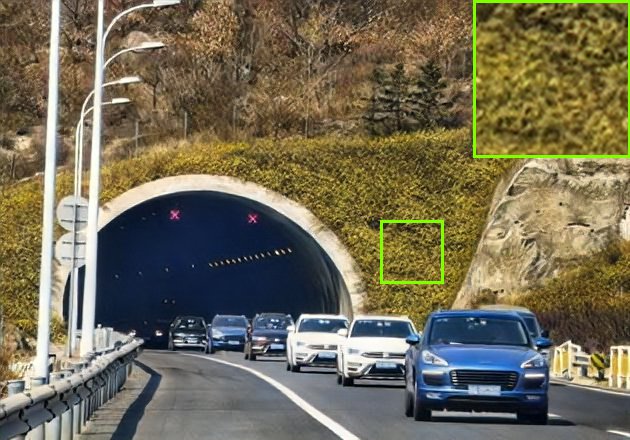}
	\end{minipage}
 	\begin{minipage}[h]{0.119\linewidth}
		\centering
		\includegraphics[width=\linewidth]{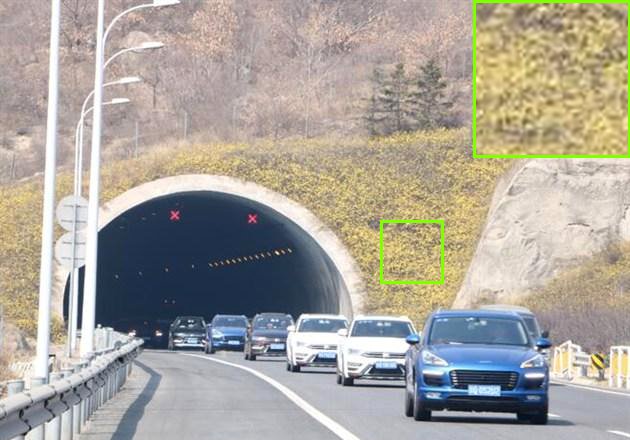}
	\end{minipage}
	\begin{minipage}[h]{0.119\linewidth}
		\centering
		\includegraphics[width=\linewidth]{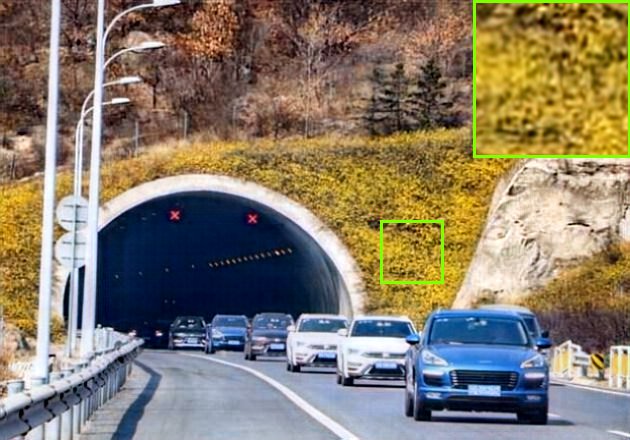}
	\end{minipage}


    \begin{minipage}[h]{0.119\linewidth}
		\centering
		\includegraphics[width=\linewidth]{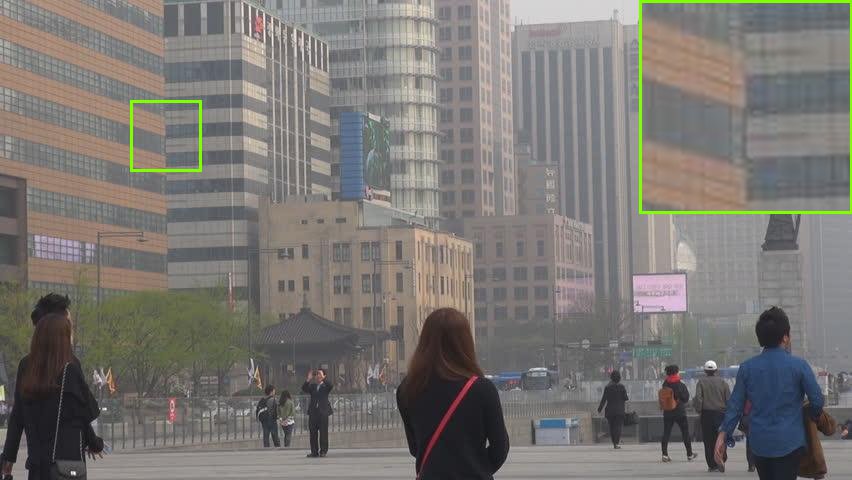}
	\end{minipage}
	\begin{minipage}[h]{0.119\linewidth}
		\centering
		\includegraphics[width=\linewidth]{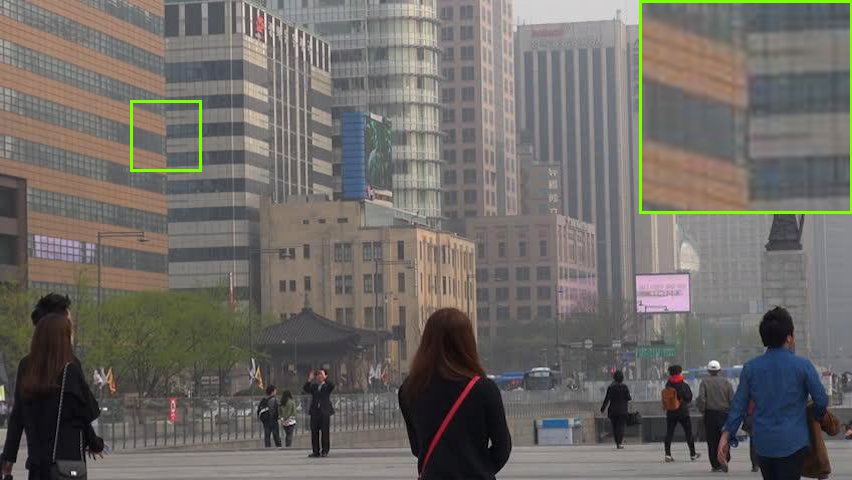}
	\end{minipage}
	\begin{minipage}[h]{0.119\linewidth}
		\centering
		\includegraphics[width=\linewidth]{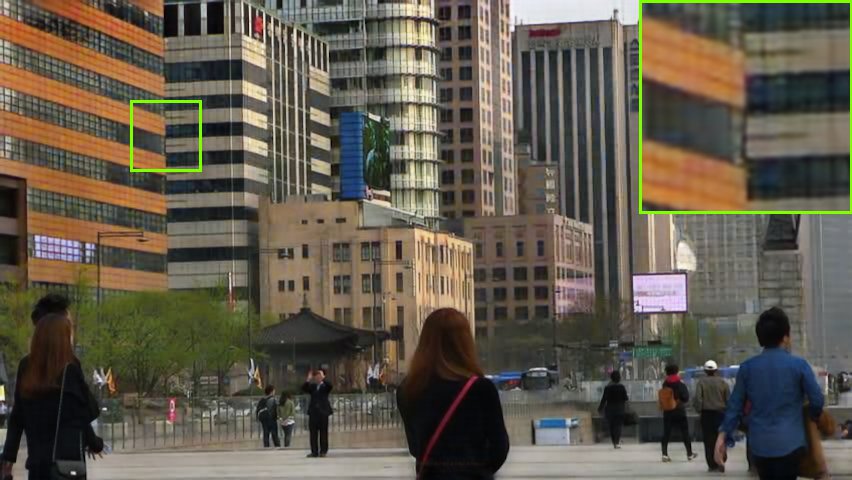}
	\end{minipage}
	\begin{minipage}[h]{0.119\linewidth}
		\centering
		\includegraphics[width=\linewidth]{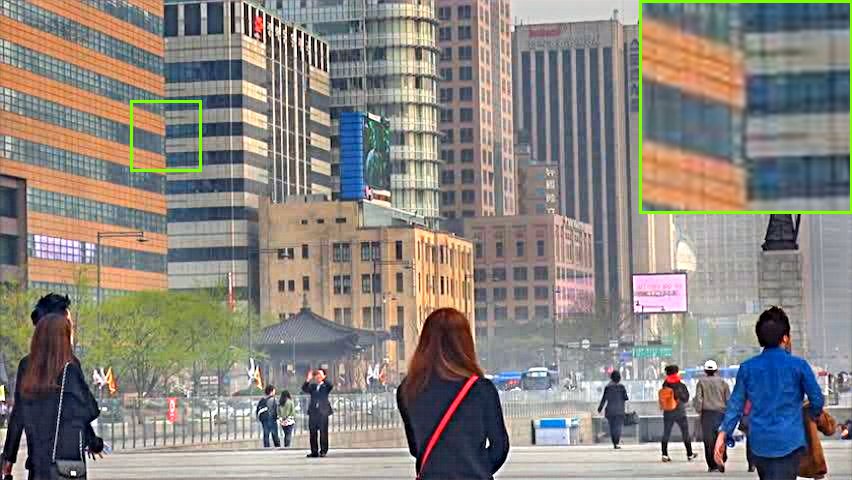}
	\end{minipage}
	\begin{minipage}[h]{0.119\linewidth}
		\centering
		\includegraphics[width=\linewidth]{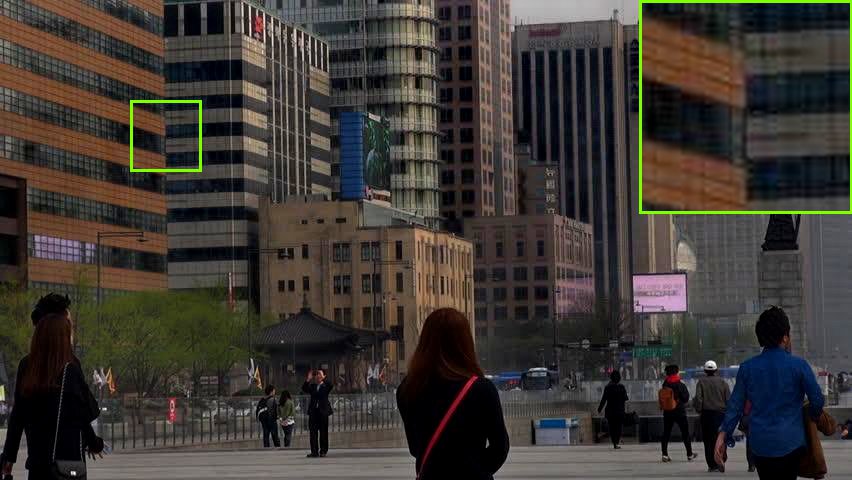}
	\end{minipage}
	\begin{minipage}[h]{0.119\linewidth}
		\centering
		\includegraphics[width=\linewidth]{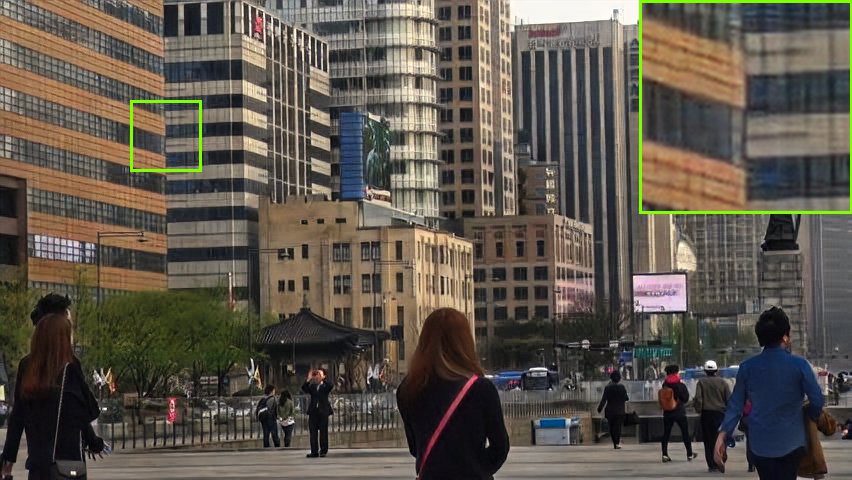}
	\end{minipage}
 	\begin{minipage}[h]{0.119\linewidth}
		\centering
		\includegraphics[width=\linewidth]{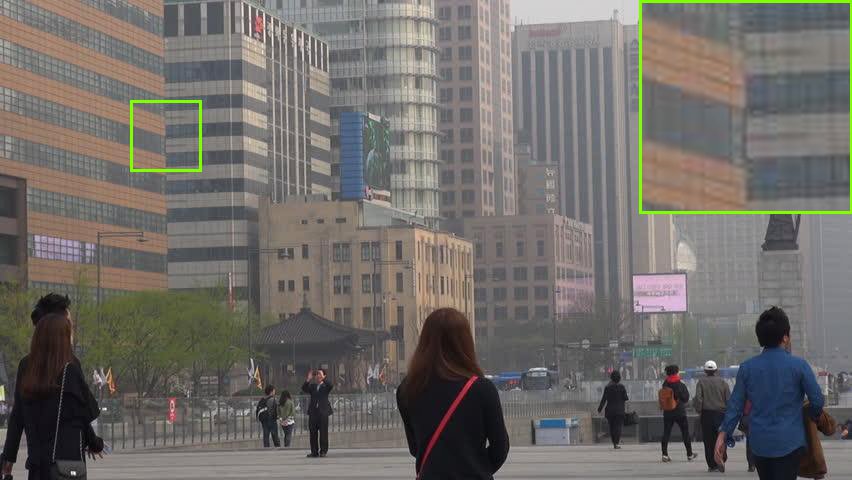}
	\end{minipage}
	\begin{minipage}[h]{0.119\linewidth}
		\centering
		\includegraphics[width=\linewidth]{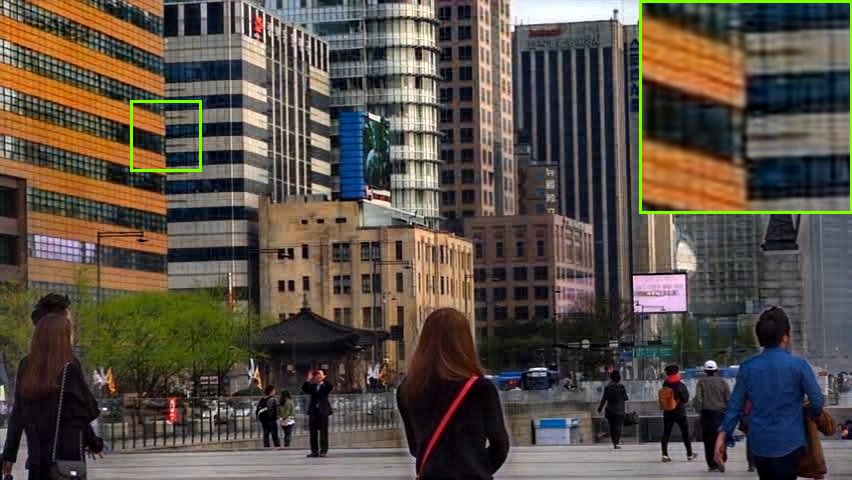}
	\end{minipage}

    \begin{minipage}[h]{0.119\linewidth}
		\centering
		\includegraphics[width=\linewidth]{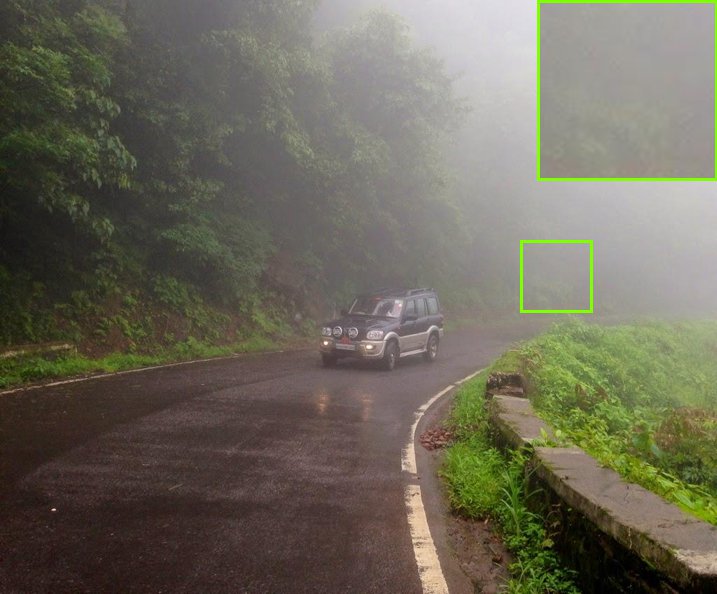}
	\end{minipage}
	\begin{minipage}[h]{0.119\linewidth}
		\centering
		\includegraphics[width=\linewidth]{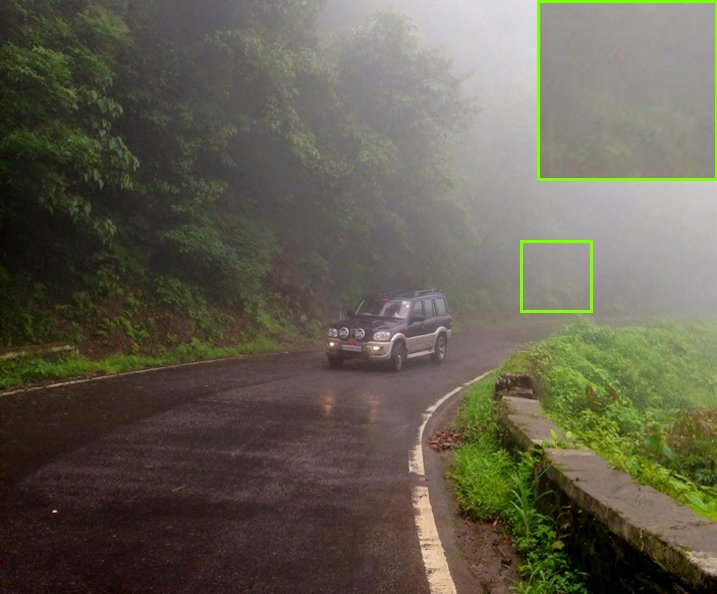}
	\end{minipage}
	\begin{minipage}[h]{0.119\linewidth}
		\centering
		\includegraphics[width=\linewidth]{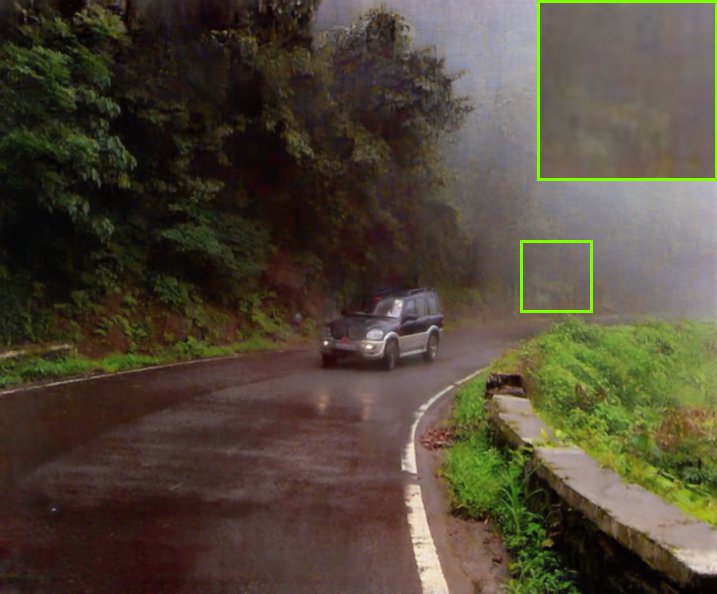}
	\end{minipage}
	\begin{minipage}[h]{0.119\linewidth}
		\centering
		\includegraphics[width=\linewidth]{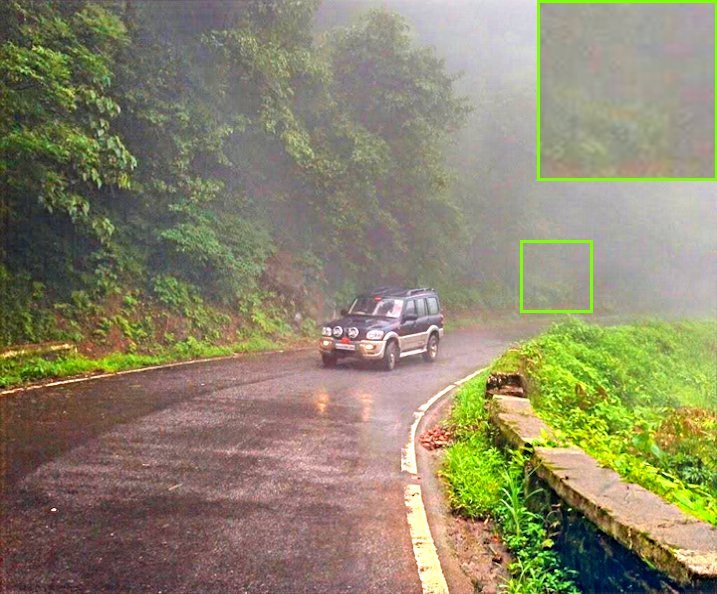}
	\end{minipage}
	\begin{minipage}[h]{0.119\linewidth}
		\centering
		\includegraphics[width=\linewidth]{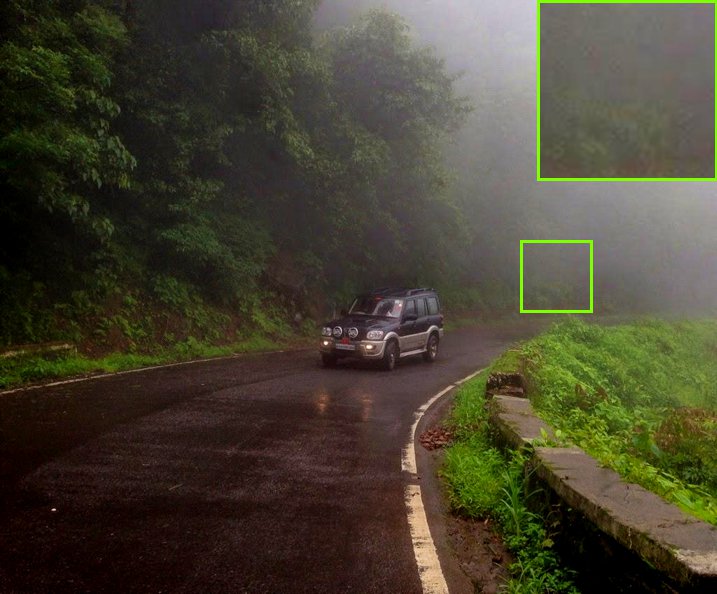}
	\end{minipage}
	\begin{minipage}[h]{0.119\linewidth}
		\centering
		\includegraphics[width=\linewidth]{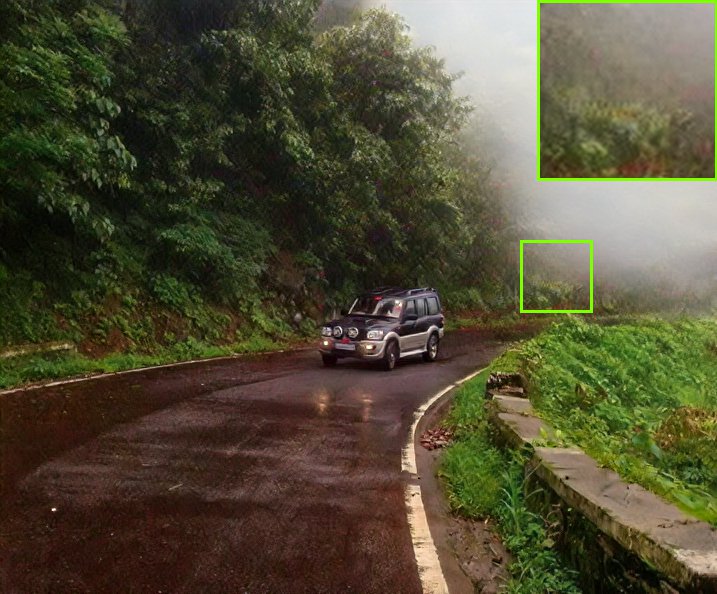}
	\end{minipage}
 	\begin{minipage}[h]{0.119\linewidth}
		\centering
		\includegraphics[width=\linewidth]{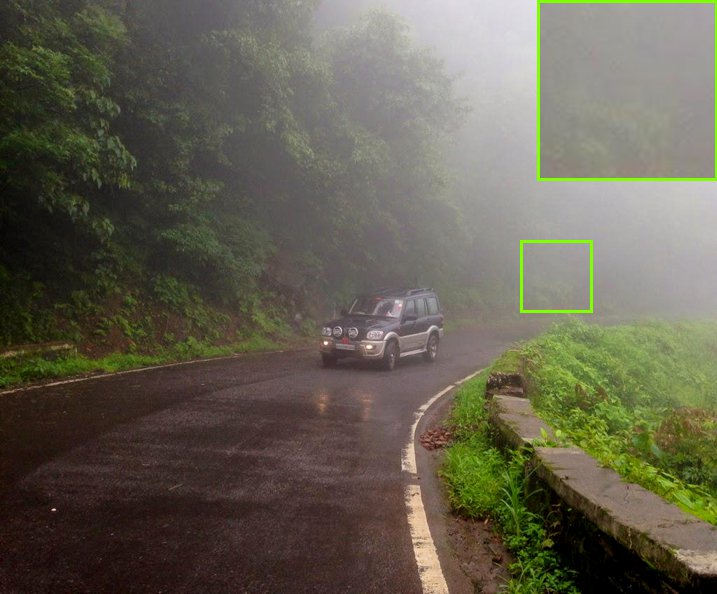}
	\end{minipage}
	\begin{minipage}[h]{0.119\linewidth}
		\centering
		\includegraphics[width=\linewidth]{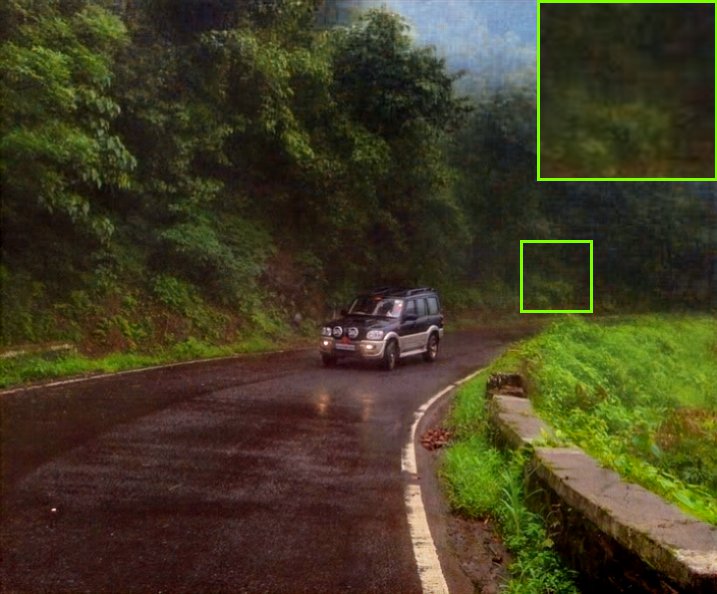}
	\end{minipage}

    \begin{minipage}[h]{0.119\linewidth}
		\centering
		\includegraphics[width=\linewidth]{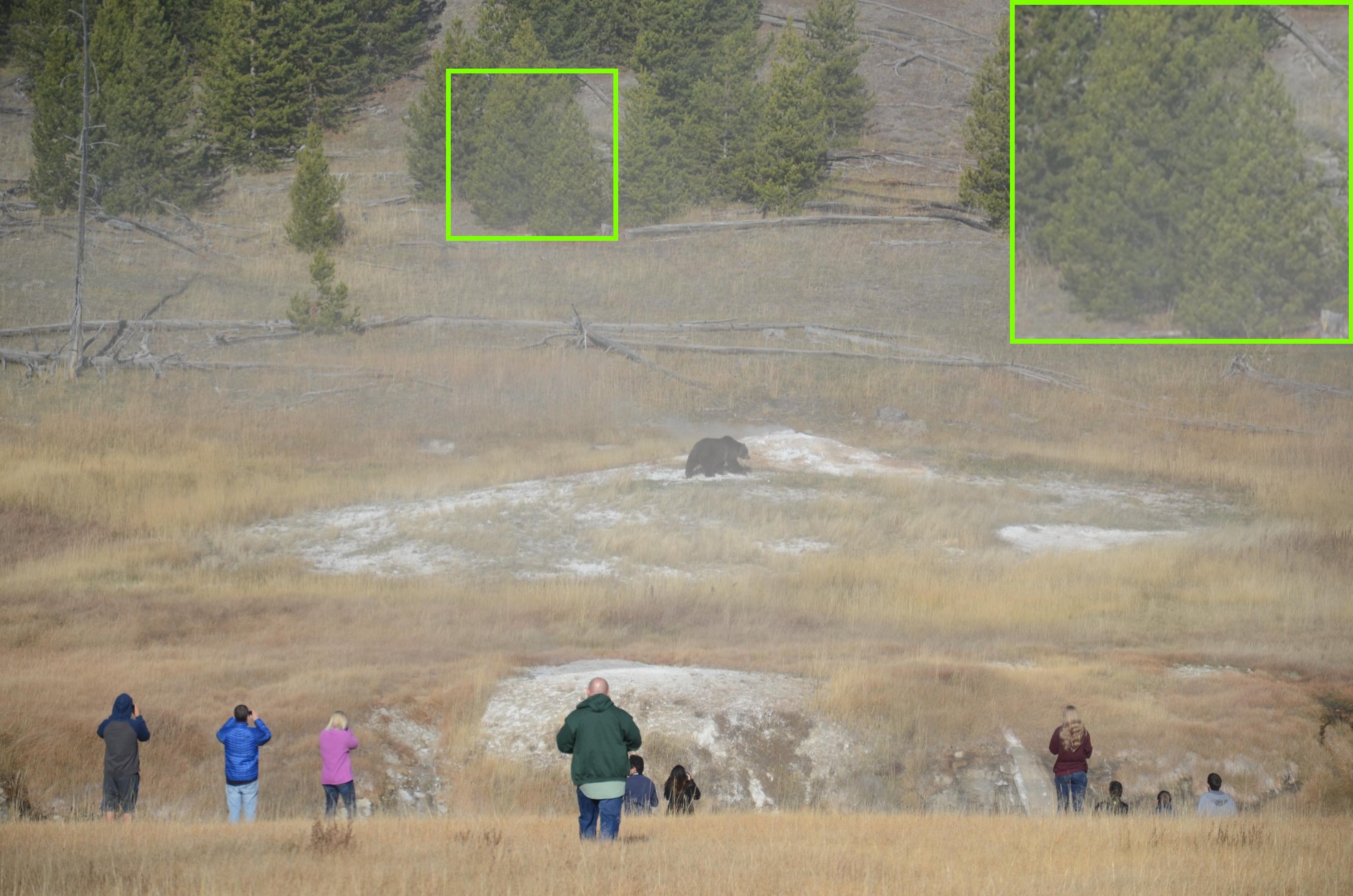}
  \small{Input}
	\end{minipage}
	\begin{minipage}[h]{0.119\linewidth}
		\centering
		\includegraphics[width=\linewidth]{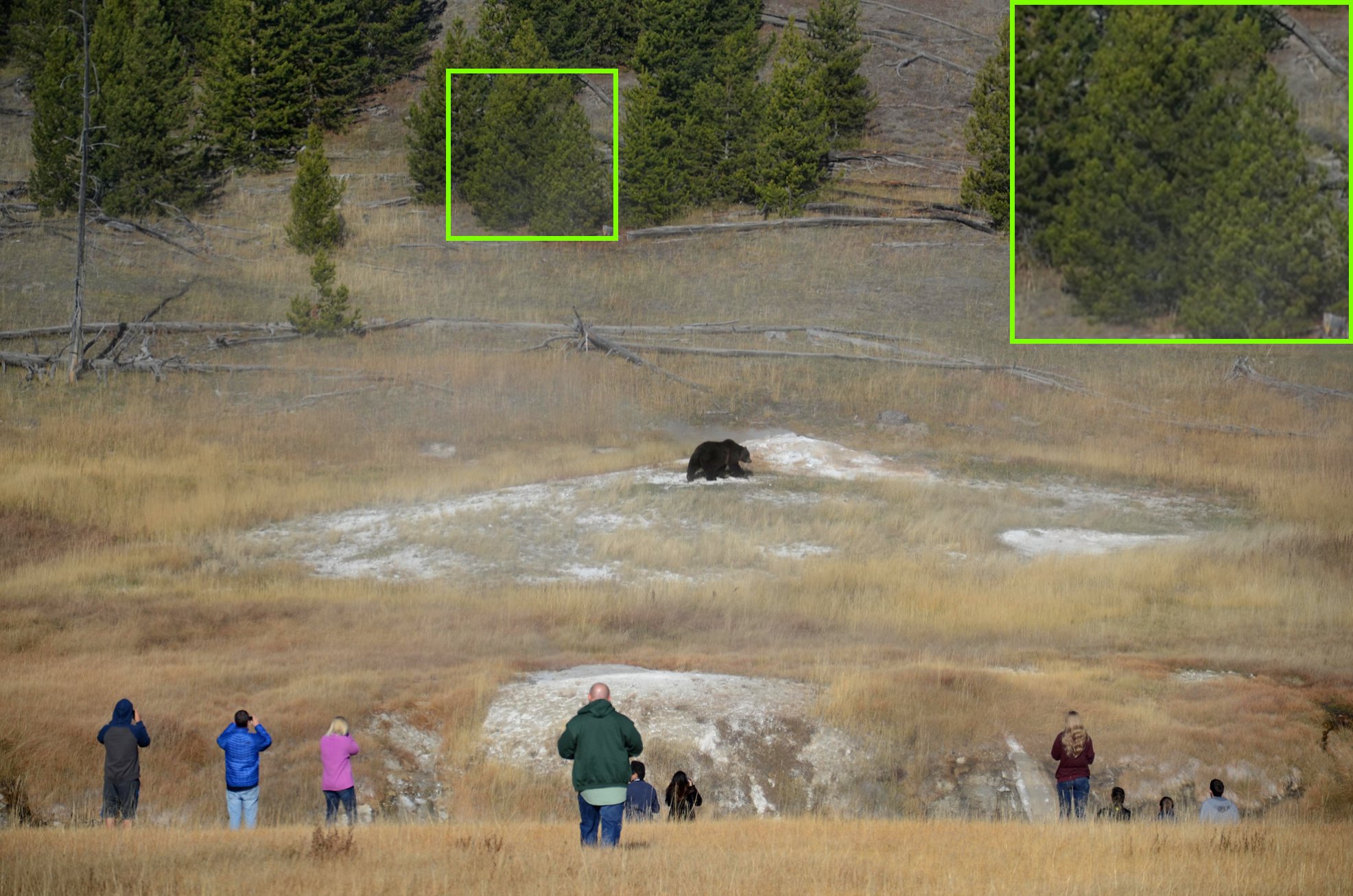}
  \small{Dehamer~\cite{Guo_2022_CVPR}}
	\end{minipage}
	\begin{minipage}[h]{0.119\linewidth}
		\centering
		\includegraphics[width=\linewidth]{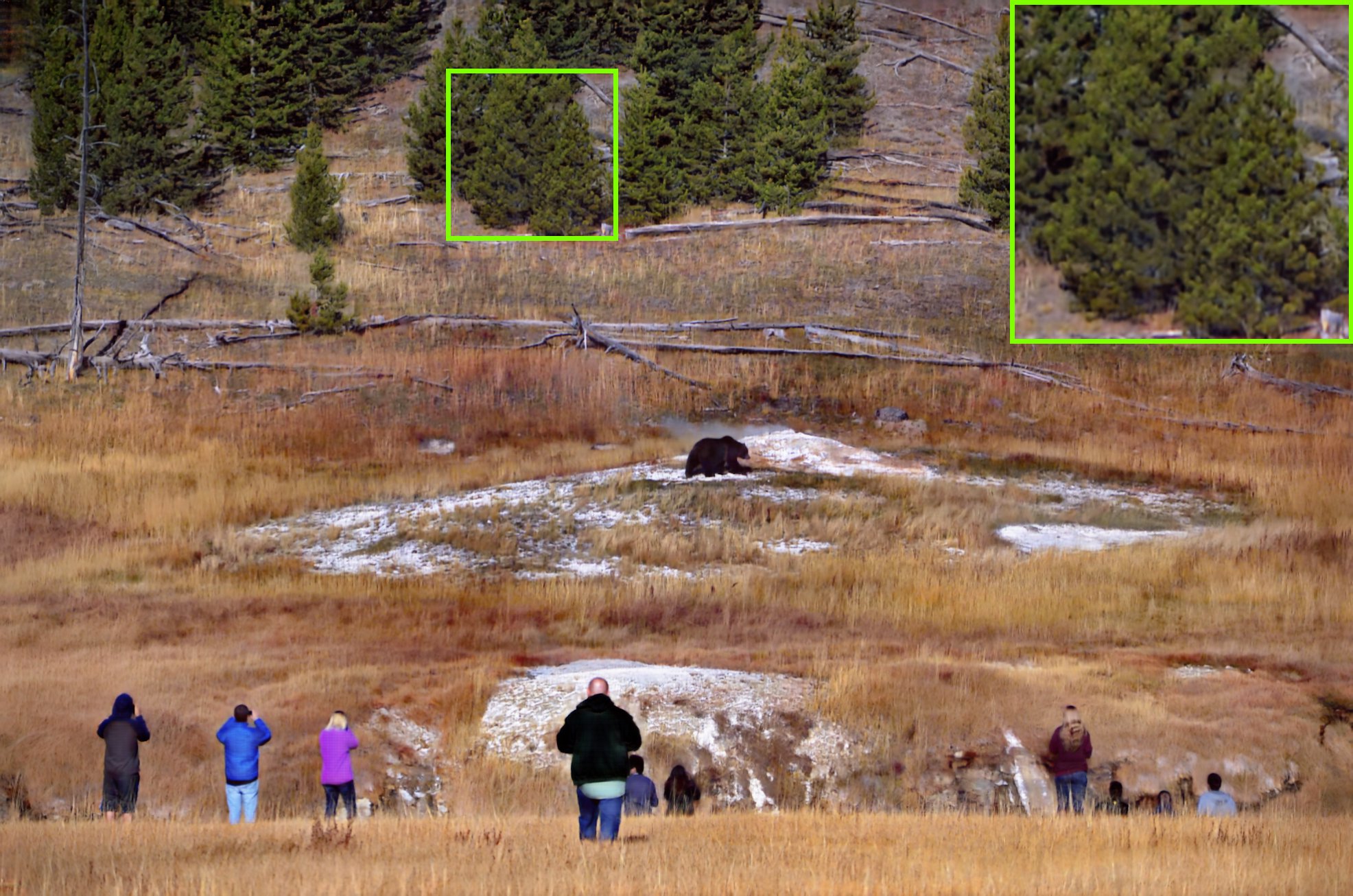}
  \small{DAD~\cite{Shao_2020_CVPR}}
	\end{minipage}
	\begin{minipage}[h]{0.119\linewidth}
		\centering
		\includegraphics[width=\linewidth]{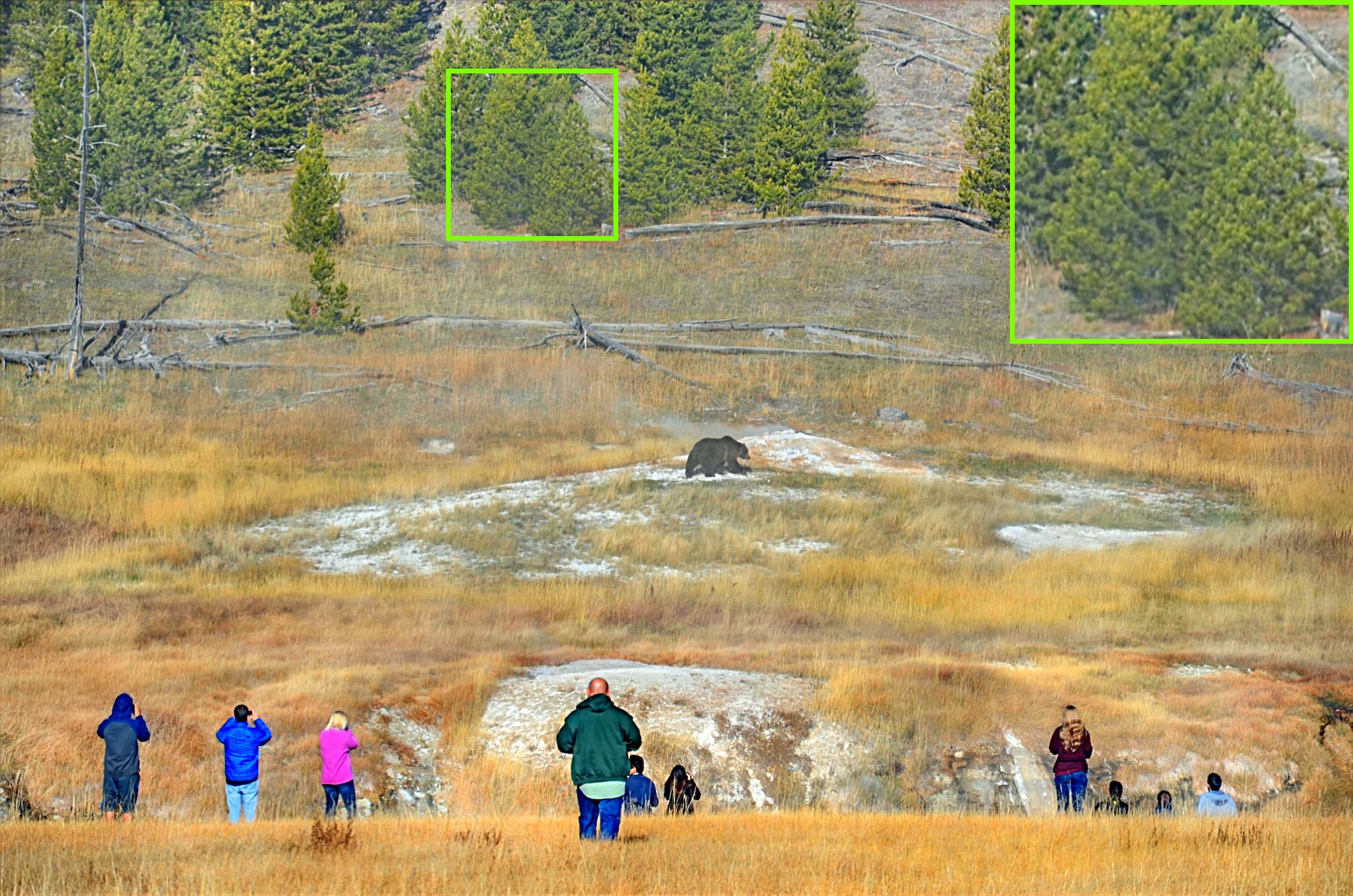}
  \small{PSD~\cite{Chen_2021_CVPR}}
	\end{minipage}
	\begin{minipage}[h]{0.119\linewidth}
		\centering
		\includegraphics[width=\linewidth]{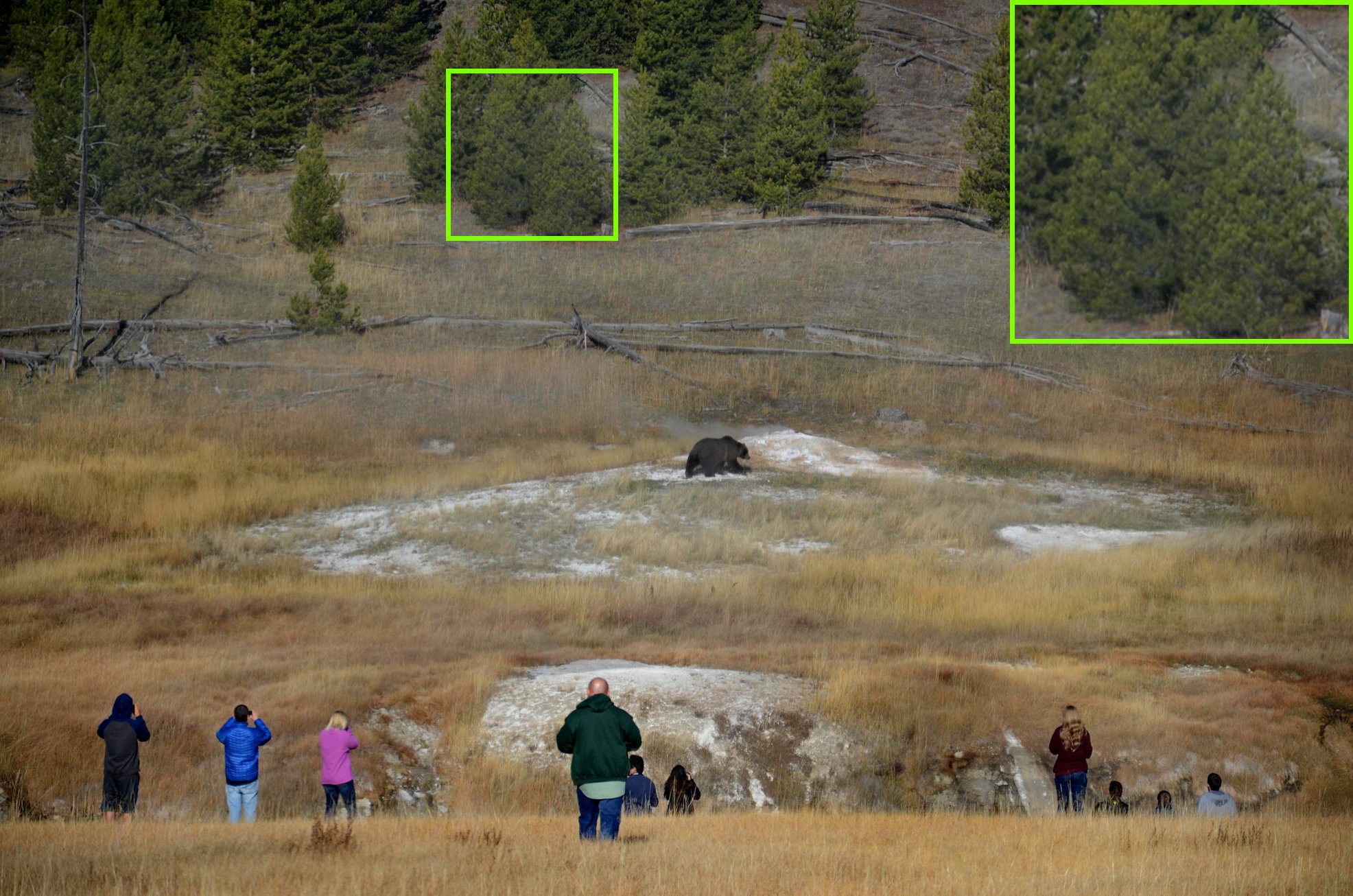}
  \small{D4~\cite{yang2022self}}
	\end{minipage}
	\begin{minipage}[h]{0.119\linewidth}
		\centering
		\includegraphics[width=\linewidth]{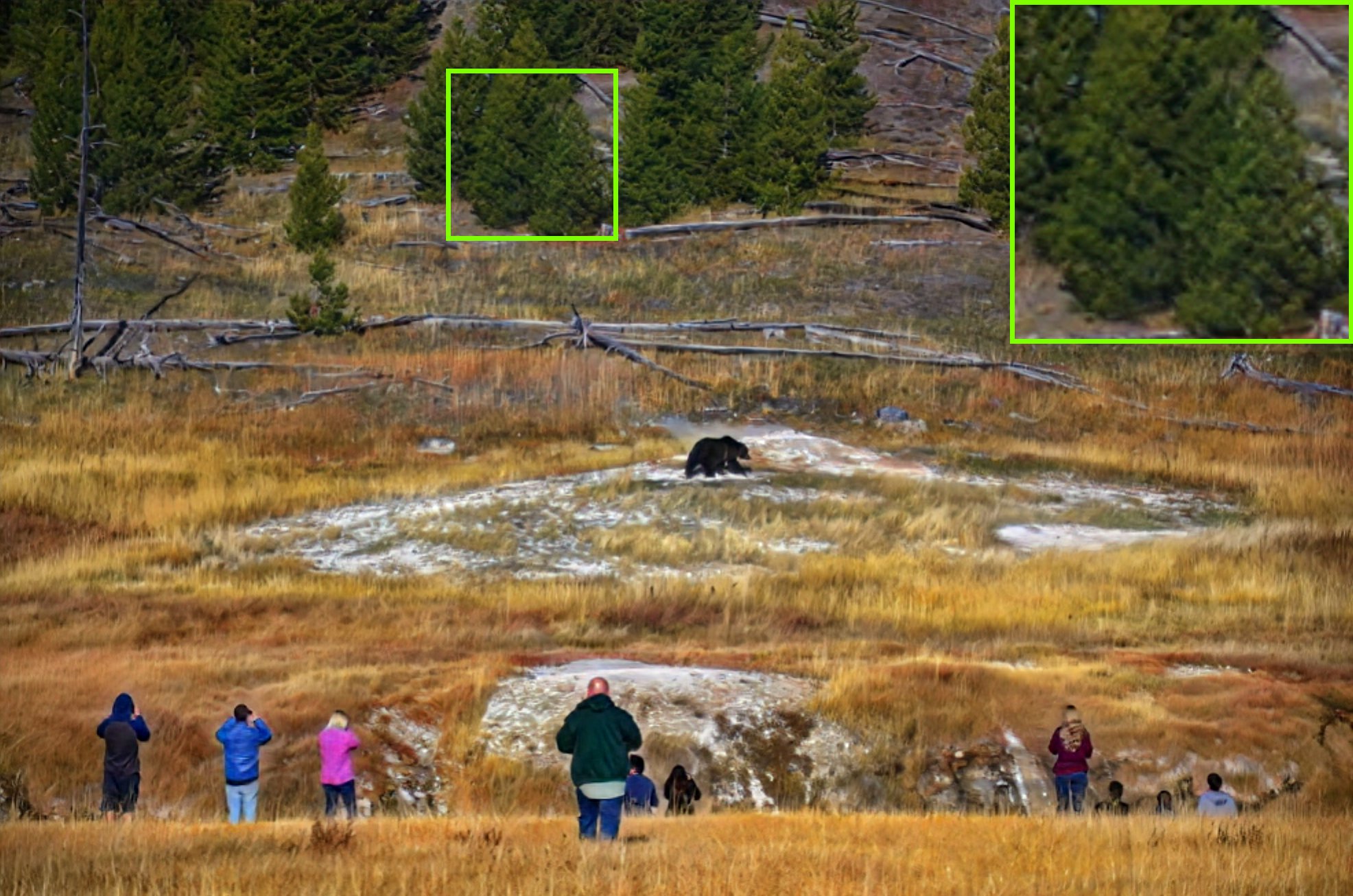}
  \small{RIDCP~\cite{Wu_2023_CVPR}}
	\end{minipage}
 	\begin{minipage}[h]{0.119\linewidth}
		\centering
		\includegraphics[width=\linewidth]{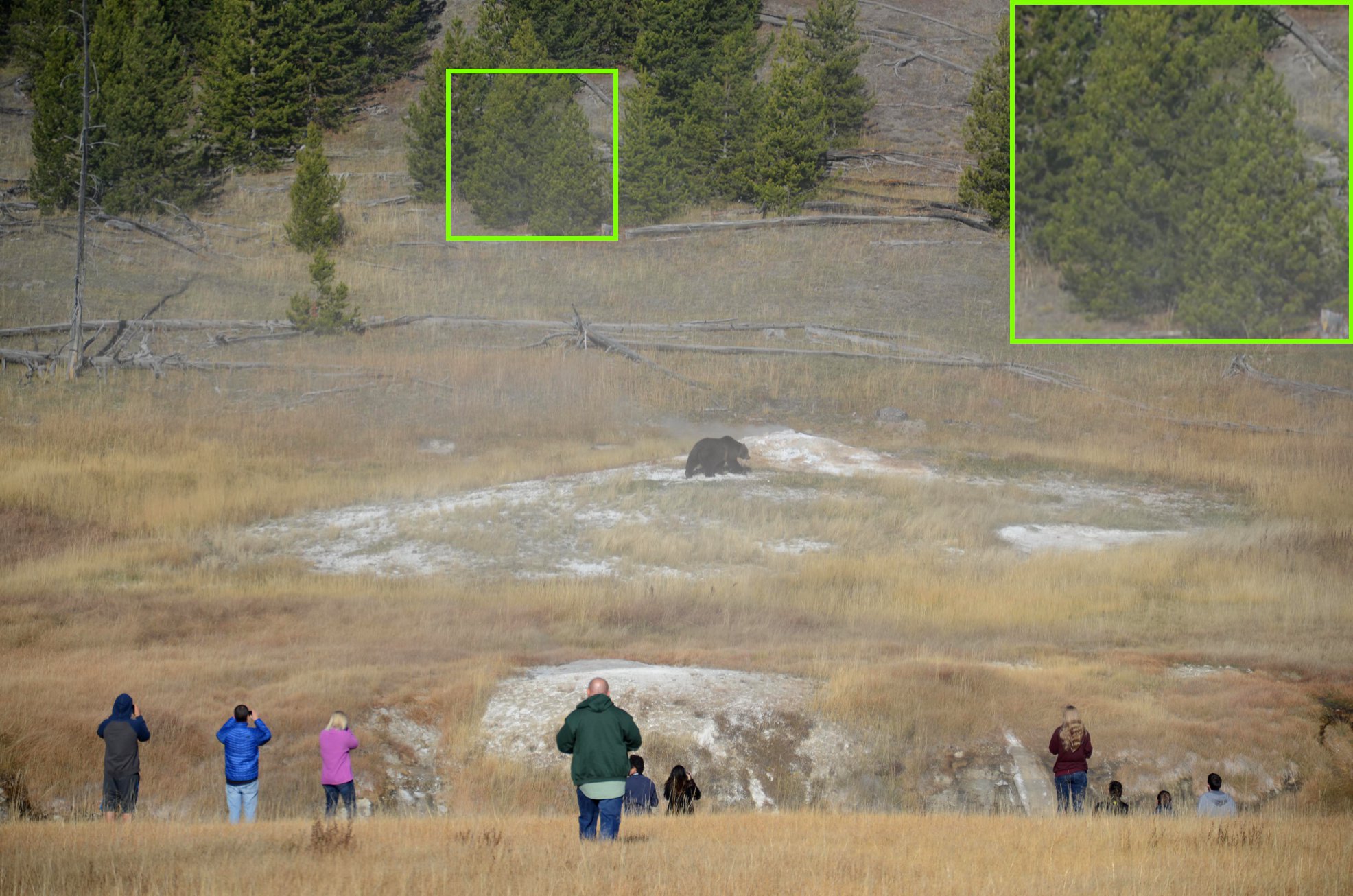}
  \small{DCMPNet~\cite{Zhang_2024_CVPR}}
	\end{minipage}
	\begin{minipage}[h]{0.119\linewidth}
		\centering
		\includegraphics[width=\linewidth]{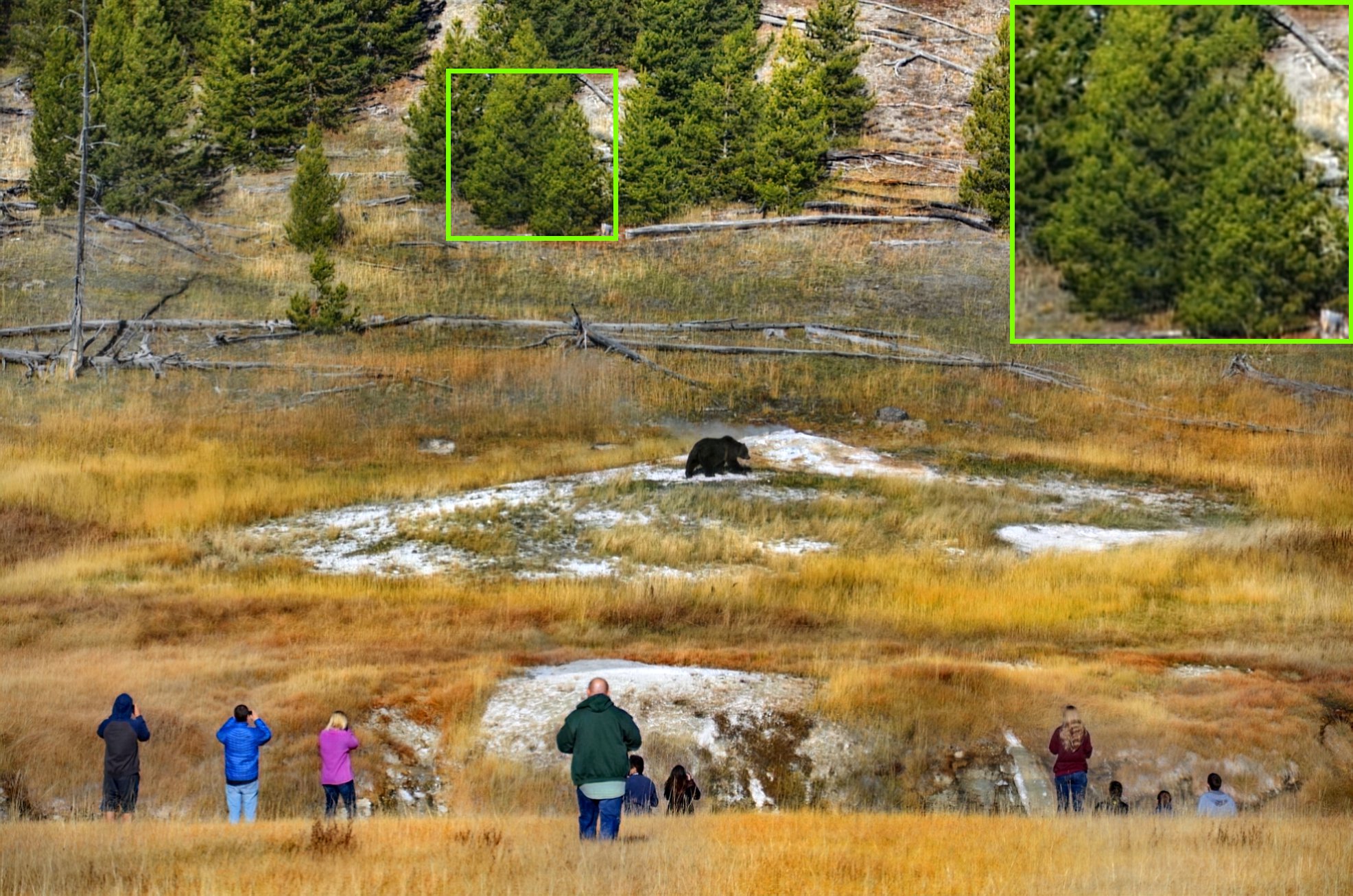}
  \small{Ours}
	\end{minipage}
    
 \centering
	\caption{Visual comparisons on RTTS~\cite{RESIDE} dataset. For better clarity, the region within the green rectangle is zoomed in and displayed in the top right corner.}
	\label{fig:qualitative_RTTS}
\end{figure*}

To mitigate the adverse impact of the concentration on sky regions, we propose a region-specific dehazing technique in addition to CLIP surgery. During fine-tuning, the non-sky regions are handled separately from the sky regions. In Fig.~\ref{fig:calibration}(d), we present the similarity maps for hazy images with the sky masked out. The CLIP model successfully identifies hazy areas in non-sky regions, providing precise guidance. Sky masking is achieved by prompting SAM model with points that have high similarity to texts describing the sky.

\subsection{Contrastive Prompt Set}
To fully leverage the capabilities of CLIP model, we construct specific prompt sets to guide the dehazing network. Each contrastive prompt set consists of a positive prompt, denoted as $T^{p}$, which captures desired properties (e.g., clean), and one or more negative prompts, denoted as $T^{n_1}$, $T^{n_2}$,  ..., $T^{n_K}$, which describe undesired properties (e.g., hazy). The text features of the positive and negative prompts are extracted by the text encoder $\Phi_{t}$ of the CLIP model. When evaluating a dehazed image $I$, the image encoder $\Phi_{i}$ of the CLIP model is employed to obtain the image features. Consequently, based on the text-image similarity in the CLIP latent space, we formulate the CLIP loss with respect to a prompt set as follows:

\begin{equation}
  \mathcal{L}_{T}(I) = \frac{e^{cos(\Phi_{i}(I), \Phi_{t}(T^p))}}{\Sigma_{j\in\{p, n_1, ..., n_K\}}e^{cos(\Phi_i(I), \Phi_{t}(T^j))}},
  \label{eq:clip_loss}
\end{equation}
where $T$$\in$$\{T_s, T_n, T_e\}$ represents the prompt set, which we introduce in the following section.


\subsection{Prompt Ensemble}
\label{sec:ensemble}
As presented in the right part of Fig.~\ref{fig:method}, to build suitable prompt sets as language guidance, we first construct a prompt template.  Positive templates include expressions like follows:

\texttt{a picture of \textless entity\textgreater~in the fog.}

\texttt{a photo of a foggy \textless entity\textgreater.}

Negative templates follow a similar structure. The sky-region dehazing prompt set $T_s$ is created by setting \texttt{sky} as \texttt{\textless entity\textgreater}. The non-sky region dehazing set $T_n$ is constructed by inserting objects like \texttt{building,} \texttt{people} as well as \texttt{scene} into the template.
In $T_s$ and $T_n$, all constructed prompts are averaged in the latent space to form one positive prompt and one negative prompt for calculating the CLIP loss.

In addition to the poor dehazing effect, the output of the pre-trained model also suffers from issues including dull colors and overall dirtiness. To mitigate this shortage, we use an additional image enhancing prompt set $T_e$  to guide the model to produce high-quality results. The design of this prompt set is flexible, and with intentional design, the fine-tuning results can even exhibit a customized style. In this paper, we apply the prompts shown in Fig.~\ref{fig:method} for the result presented.

\subsection{Synthetic-to-Real Adaptation}

With a pre-trained dehazing network and three contrastive prompt sets, we are able to generalize the model from synthetic to real-world domain. Given an unlabelled real-world hazy image $I$, sky region $I_s$ and non-sky regions $I_n$ are separated. The CLIP guidance loss is formulated as:
\begin{equation}
  \mathcal{L}_{c}(I) = \mathcal{L}_{T_s}(\mathcal{M}(I_s)) + \mathcal{L}_{T_n}(\mathcal{M}(I_n))
  + \lambda_1 \cdot \mathcal{L}_{T_e}(\mathcal{M}(I)),
\end{equation}
where $\lambda_1$ represents the tradeoff weight, $L_T$ is defined in Eq.~\ref{eq:clip_loss}, and $\mathcal{M}$ denotes the dehazing network.

To avoid catastrophic forgetting during fine-tuning, following \cite{CLIP-LIT}, we implement a fidelity loss to constrain the dehazing result, with $\alpha_l$ representing the tradeoff weight of the $l$-th layer in the pre-trained CLIP image encoder $\Phi_{i}$:
\begin{equation}
\mathcal{L}_{f}(I) = \Sigma^{4}_{l=0} \alpha_l\cdot \Vert \Phi^l_{i}(\mathcal{M}(I)) - \Phi^l_{i}(I) \Vert_2.
\end{equation}

Eventually, the overall loss function $\mathcal{L}$ in fine-tuning is defined with respect to a tradeoff weight $\lambda_2$:
\begin{equation}
\mathcal{L}(I) = \mathcal{L}_{c}(I) + \lambda_2 \cdot\mathcal{L}_{f}(I).
\end{equation}

\section{Experiments}

\subsection{Experimental Settings}
\textbf{Dataset.} For pre-training, we used synthetic data from RIDCP~\cite{Wu_2023_CVPR}. For fine-tuning, we selected the URHI split from the RESIDE~\cite{RESIDE} dataset. Following previous works~\cite{Wu_2023_CVPR}, we tested our method on the RTTS split from the RESIDE dataset, which contains over 4,000 real-world hazy images with diverse scenes and haze patterns.
\begin{table}[t]
  \centering
    \caption{Quantitative comparison on RTTS dataset~\cite{RESIDE}. The best is in \textcolor[rgb]{ 1,  0,  0}{\textbf{red}} while the second is in \textcolor[rgb]{ 0,  0,  1}{\textbf{blue}}.}
  \label{tab:NR}%
  \scalebox{0.95}{
    \begin{tabular}{c|ccccc}
    \toprule
    Method & FADE↓ & RRPD↑ & BRISQUE↓ & NIMA↑ & MOS↑\\
    \midrule
    Hazy image & 2.484 & - & 37.011 & 4.3250 & - \\
    Dehamer~\cite{Guo_2022_CVPR} & 1.895 & 51.59 & 33.866 & 3.8663 & 2.78 \\
    DAD~\cite{Shao_2020_CVPR}   & 1.130 & 179.29 & 32.727 & 4.0055 & 3.13\\
    PSD~\cite{Chen_2021_CVPR}   & \textcolor[rgb]{0,0,1}{\textbf{0.920}} & 173.94 & 25.239 & 4.3459 & 3.20\\
    D4~\cite{yang2022self}    & 1.358 & 166.56 & 33.206 & 3.7239 & 2.48\\
    RIDCP~\cite{Wu_2023_CVPR} & 0.944 & \textcolor[rgb]{0,0,1}{\textbf{240.42}} &\textcolor[rgb]{0,0,1}{\textbf{18.782}} & 4.4267 & \textcolor[rgb]{0,0,1}{\textbf{3.57}} \\
    DCMPNet~\cite{Zhang_2024_CVPR}  & 1.921 & 45.37 & 32.520 & \textcolor[rgb]{0,0,1}{\textbf{4.4351}} & 2.85\\
    HazeCLIP & \textcolor[rgb]{1,0,0}{\textbf{0.638}} & \textcolor[rgb]{1,0,0}{\textbf{260.35}} & \textcolor[rgb]{1,0,0}{\textbf{18.567}} & \textcolor[rgb]{1,0,0}{\textbf{4.5510}} & \textcolor[rgb]{1,0,0}{\textbf{3.60}}\\
    \bottomrule
    \end{tabular}}

\end{table}%

\textbf{Implementation Details.} 
HazeCLIP is implemented using PyTorch 1.13, and all experiments are conducted on a single NVIDIA RTX 4090 GPU. Unless otherwise specified, the dehazing network is set to be MSBDN~\cite{Dong_2020_CVPR}. During pre-training, the network was trained for 200 epochs using the Lion optimizer. The initial learning rate was set to $3\times10^{-5}$ with a cosine annealing scheduler. The network was later fine-tuned for 15 epochs. We empirically set $\lambda_1=0.5$ and $\lambda_2=0.1$. The weighting factors, $\alpha_0$ to $\alpha_3$, are set to 1.0, while $\alpha_4$ is set to 0.5.

\subsection{Comparison with State-of-the-Art Methods}
We compared the performance of HazeCLIP with several state-of-the-art dehazing approaches: Dehamer~\cite{Guo_2022_CVPR}, DAD~\cite{Shao_2020_CVPR}, PSD~\cite{Chen_2021_CVPR}, D4~\cite{yang2022self}, RIDCP~\cite{Wu_2023_CVPR}, and DCMPNet~\cite{Zhang_2024_CVPR}. It's worth noting that since HazeCLIP doesn't introduce any new parameters, the computational cost during inference remains the same as that of the backbone network.

\textbf{Visual Quality.} 
We evaluated the visual quality of HazeCLIP on real-world hazy images from the RTTS dataset. As shown in Fig.~\ref{fig:qualitative_RTTS}, Dehamer and DCMPNet struggle with effective real-world haze removal. D4's results exhibit an unpleasantly dark tone, while PSD produces unrealistically bright outcomes. DAD and RIDCP leave residual haze in some images. In contrast, the proposed method demonstrates superior overall performance, effectively addressing the shortcomings of the other approaches.

\textbf{Image Quality Assessment Metrics.}                 
For dehazing performance comparison, we used no-reference evaluator FADE~\cite{fade} and reduced-reference evaluator RRPD~\cite{rrpd}. We also included two widely-used no-reference image quality assessment metrics BRISQUE~\cite{brisque} and NIMA~\cite{nima}. Additionally, we conducted a user study that invited 22 subjects to obtain the subjective Mean Opinion Score (MOS). The results, reported in Table~\ref{tab:NR}, show that HazeCLIP achieved the best performance in all metrics. Notably, HazeCLIP demonstrated a 30.7\% improvement in FADE, underscoring its superiority in real-world dehazing.

\begin{table}[t]
  \centering
  \caption{Quantitative comparisons among the three settings of Ablation Study and the full version of HazeCLIP. [Key: \textbf{Best}]}
  \label{tab:ablation}%
  \scalebox{0.95}{
    \begin{tabular}{c|cccc}
    \toprule
    Metric & setting (a) & setting (b) & setting (c) & HazeCLIP \\
    \midrule
    FADE ↓ & 1.091 & 0.857 & 0.695 & \textbf{0.638} \\
    RRPD ↑ & 200.42 & 245.51 & 258.39 & \textbf{260.35} \\
    BRISQUE ↓ & 27.309 & 20.499 & 22.376 & \textbf{18.567} \\
    NIMA ↑ & 4.3961 & 4.4587 & 4.3965 & \textbf{4.5510} \\
    \bottomrule
    \end{tabular}}
\end{table}%

\subsection{Ablation Study}

To verify the effectiveness of each key component, we conducted a series of ablation experiments with the following variants to the full framework: (a) without HazeCLIP adaptation; (b) without the region-specific dehazing technique (i.e., a single dehazing prompt set applies to the whole image); and (c) without the enhancing prompt set. As shown in Table~\ref{tab:ablation}, quantitative metrics demonstrate the necessity of the full framework. The pre-trained model performs poorly without HazeCLIP adaptation. Without the region-specific dehazing technique, the dehazing effect is impaired, as evidenced by FADE and RRPD. The overall image quality is reduced in the absence of the enhancing set.

\begin{table}[t]
  \centering
    \caption{Quantitative comparisons of pre-trained and fine-tuned versions of different networks. [Key: \textbf{Best}]}
    \scalebox{0.95}{
  \label{tab:gen}%
    \begin{tabular}{c|cccc}
    \toprule
    Method & FADE↓ & RRPD↑ & BRISQUE↓ & NIMA↑ \\
    \midrule
    GDN~\cite{GDN} & 1.470 & 176.20 & 29.448 & 4.2502 \\
    GDN+HazeCLIP & \textbf{0.976} & \textbf{219.22} & \textbf{21.352} & \textbf{4.3252} \\
    \midrule
    FFANet~\cite{ffa} & 1.153 & 190.23 & 26.233 & 4.2817 \\
    FFANet+HazeCLIP & \textbf{0.913} & \textbf{235.61} & \textbf{19.522} & \textbf{4.4019} \\
    \midrule
    MSBDN~\cite{Dong_2020_CVPR} & 1.091 & 200.42 &  27.309 & 4.3961 \\
        MSBDN+HazeCLIP & \textbf{0.638} & \textbf{260.35} & \textbf{18.567} & \textbf{4.5510} \\

    \bottomrule
    \end{tabular}}
\end{table}%

\subsection{Framework Generalization}
 To confirm that HazeCLIP is a general framework capable of fine-tuning many image dehazing networks, we applied it to two additional popular models: FFANet~\cite{ffa} and GDN~\cite{GDN}. We compared the performance of these pre-trained models with the models fine-tuned using our HazeCLIP framework. As shown in Table~\ref{tab:gen}, HazeCLIP consistently improved the performance of the three networks across all metrics.

\section{Conclusion}

In this paper, we introduce HazeCLIP, an innovative adaptation framework designed to generalize dehazing networks pre-trained on synthetic data to real-world applications. By employing the region-specific dehazing technique and specially designed prompt sets, HazeCLIP accurately detect hazy regions and guide the fine-tuning process. Both subjective evaluations and quantitative metrics demonstrate the superior performance of the proposed approach. Additionally, HazeCLIP is compatible with various dehazing networks, showing its practical value. We hope that HazeCLIP will inspire new directions in the integration of vision-language models with broader image restoration efforts.

\textbf{Limitations.} During development of HazeCLIP, we also identify two limitations. First, contrastive prompt sets can be constructed through a learned or more systematic approach.
Additionally, robust reduced-reference and no-reference evaluation metrics for real-world image dehazing are lacking.

\vfill\pagebreak

\balance
\bibliographystyle{IEEEtran}
\bibliography{main}

\end{document}